\documentclass[10pt,twocolumn,letterpaper]{article}

\usepackage{3dv}
\usepackage{times}
\usepackage{epsfig}
\usepackage{graphicx}
\graphicspath{ {resources/} }
\usepackage{amsmath}
\usepackage{amssymb}
\usepackage{bm} 
\usepackage{mathtools} 
\usepackage{dsfont} 
\usepackage[toc,page]{appendix} 
\usepackage{subcaption}
\usepackage{tabularx}
\usepackage{float}
\usepackage{placeins}
\usepackage{gensymb} 
\usepackage[pagebackref=true,breaklinks=true,colorlinks,bookmarks=false]{hyperref}

\usepackage{multirow}

\newcommand{\para}[1]{\textbf{#1}} 

\DeclareMathOperator*{\argmax}{arg\,max}
\DeclareMathOperator*{\argmin}{arg\,min}

\DeclarePairedDelimiter{\norm}{\lVert}{\rVert}

\newcommand\scalemath[2]{\scalebox{#1}{\mbox{\ensuremath{\displaystyle #2}}}} 

\makeatletter
\def\blfootnote{\gdef\@thefnmark{}\@footnotetext}
\makeatother

\threedvfinalcopy 

\def\threedvPaperID{} 

\begin{document}

\title{Motion Capture from Pan-Tilt Cameras with Unknown Orientation}

\author{
Roman Bachmann$^1$
\and
J\"org Sp\"orri$^2$
\and
Pascal Fua$^1$
\and
Helge Rhodin$^{1,3}$\vspace{0.1cm}
\and
$^1$EPFL, Lausanne, Switzerland\\
$^2$Balgrist University Hospital, University of Zurich, Zurich, Switzerland\\
$^3$The University of British Columbia, Vancouver, Canada\\
{\tt\small firstname.lastname@epfl.ch}
}

\maketitle

\begin{abstract}
In sports, such as alpine skiing, coaches would like to know the speed and various biomechanical variables of their athletes and competitors. Existing methods use either body-worn sensors, which are cumbersome to setup, or manual image annotation, which is time consuming.
We propose a method for estimating an athlete's global 3D position and articulated pose using multiple cameras. By contrast to classical markerless motion capture solutions, we allow cameras to rotate freely so that large capture volumes can be covered. 
In a first step, tight crops around the skier are predicted and fed to a 2D pose estimator network. The 3D pose is then reconstructed using a bundle adjustment method. Key to our solution is the rotation estimation of Pan-Tilt cameras in a joint optimization with the athlete pose and conditioning on relative background motion computed with feature tracking. Furthermore, we created a new alpine skiing dataset and annotated it with 2D pose labels, to overcome shortcomings of existing ones.
Our method estimates accurate global 3D poses from images only and provides coaches with an automatic and fast tool for measuring and improving an athlete's performance.
\end{abstract}

\section{Introduction}
	In many sports, like alpine skiing, coaches would like to know performance metrics such as Center of Mass, speed and various biomechanical variables at every point in time, giving them accurate feedback about potential increases or losses in speed and precision. This can be used to enhance the athlete's performance by comparing athletes and finding optimal motion trajectories. Existing methods, like optical barriers in skiing, only offer average speeds within segments, while other methods using Inertial Measurement Unit and/or GPS sensors \cite{AlpineSkiingKinematics, Gilgien15, Gilgien14, IMUPerformanceAnalysis, GNSSPostureEstimation} are cumbersome to wear. Using motion capture suits is also not feasible in high-speed settings with large capture volumes.

	Recent deep-learning-based monocular human pose estimation methods are able to reconstruct articulated 3D pose from moving cameras~\cite{Martinez17,Mehta17a,Pavlakos16,Tome17,Popa17,Moreno17,Rogez17,Pavlakos17,Zhou17d,Tekin17a,Sun17,Zanfir18a}, however, only relative to the camera pose and without accurate scale and depth information \cite{gunel2018face}. Such relative poses contain no information on the athletes global position and speed, the unquestionably most important metric for racing sports. We therefore have the goal of estimating an athlete's global 3D pose at every point in time using just video frames from multiple cameras arranged around the track. One way to get those poses is to manually annotate every frame. This manual annotation is however very tedious and time-consuming, so instead we chose to train a pose estimation network to predict 2D joint locations without the athletes needing to wear markers. Normally, pose estimation algorithms are only trained on human pose databases which don't feature motions of particular sports. We focus on alpine skiing, where no suitable dataset exists. Existing alpine skiing datasets~\cite{SpoerriInjuryPrevention, AlpineSkiingKinematics, RhodinUnsupervisedSki} are very limited in the number of athletes and locations that they feature, making methods trained on them not generalize well. To remedy those problems we created a new alpine skiing dataset, containing 1982 manually annotated frames from various recordings and in diverse weather conditions.

	To go from videos to global articulated 3D poses, we propose the following multi-stage approach. Because athletes are often very small in the captured images, we first train a network to predict a tight bounding box around them. Those crops are then given to the pose estimation network that was mainly trained on the new skiing dataset. The 2D detections from all cameras are then combined in a bundle adjustment approach to reconstruct the global 3D pose. Key to our solution was the expression of the athletes's motion in terms of a discrete cosine basis, which enforces smoothness constraints explicitly, and tracking features on the static background as an additional cue for constraining the camera motion. This strategy is closely related to panorama stitching and structure from motion and more general than related methods that utilize known line markings on sports fields \cite{SportCameraCalib, PTZCameraCalibSport, PTZCameraCalibMovingPeople}.
	We tried off-the-shelf structure from motion methods, however, these fail on skiing footage due to the large zoom, large distance and view angle between cameras, and lack of discriminative patterns on the ski slope. Moreover, optimizing camera motion freely as in \cite{Takahashi} lead to underconstrained systems of equations and diverging behaviour. 
    We evaluate the performance of taking fully calibrated cameras and our method for estimating the camera rotations with varying number of cameras. Our results are significantly more accurate across various biomechanical variables when compared to using the monocular reconstruction of~\cite{RhodinUnsupervisedSki}.

\section{Related work}
\label{sec:related_work}
As this paper builds upon work in the field of 3D human pose estimation, we outline in the following, the most important advances in this area and also explain our need for a new task-specific skiing dataset.

\para{Global 3D human pose estimation.}
Using at least two cameras from different perspectives, it is possible to obtain a global 3D pose estimate and potential ambiguities in scale can be resolved \cite{Moeslund01a,Moeslund06}. It is now common to estimate 2D pose with deep neural networks \cite{OpenPose,Insafutdinov16a} and infer skeleton pose with model-based optimization \cite{Rhodin16b,Huang18}.
More recently, Pavlakos \etal\cite{Pavlakos17} propose to extend pictorial structure models by taking CNN generated 2D heatmaps and resolving the 3D structure in a quantized grid by maximizing a likelihood term explaining the 2D detections. This line of work requires known camera pose and intrinsic parameters.

Puwein \etal\cite{JointCameraPoseEst} jointly estimate a 3D human pose and the position and orientation of several fixed wide-baseline cameras using a bundle adjustment method that minimizes an energy function comprising reprojection errors, a smoothness term and optical flow consistency between the motion of the estimated kinematic structure and the videos.
Similarly, Elhayek \etal\cite{OutdoorSimultPoseCamOptim} estimate both pose and camera locations simultaneously, with the difference that some cameras are fixed, while a small subset can freely move. They minimize an energy function containing a negative likelihood term describing the similarity of the model parameters to the measured data, as well as smoothness terms for both the human pose and the cameras.

More difficult is the reconstruction from moving cameras with totally unknown orientation. Several papers~\cite{SportCameraCalib, PTZCameraCalibSport, PTZCameraCalibMovingPeople} leverage common line markings of sports fields as known reference points for pan-tilt-zoom (PTZ) camera calibration. Those methods can leverage the geometric constraints that games like football are played on a two-dimensional surface with a limited spatial extent, but don't generalize to sports with unconstrained environments, such as ski racing.

Using multiple hand-held and unsynchronized cameras, Hasler \etal~\cite{HaslerMarkerless} first construct a global model of the environment, then synchronize the cameras using sound, and finally use a silhouette-based approach to find a 3D mesh of the subject. In our case, reconstruction on very zoomed in, fast moving motion would not yield enough overlap to employ the structure from motion approach described.

Most closely related to ours is the approach of Takahashi \etal\cite{Takahashi}, using multiple unsynchronized and uncalibrated cameras. They use a bundle adjustment method that leverages the limb lengths as priors on the human body and takes into account that the 2D pose estimations contain some amount of error. However, using their approach to freely optimize camera rotations failed in our large capture volume with very fast moving athletes.

\para{Ski datasets.}
While there exists extensive datasets for human poses in various settings like the MPII Human Pose~\cite{MPIIHumanPoseDataset} dataset or the Human3.6M~\cite{Human3.6M} dataset, they feature only very few, if any, skiing images of amateurs and lack annotation of the skis and poles. Professional athletes in a racing scenario are even more rare, which would make accurate inference impossible in those cases. To train or refine a 2D detector, we have therefore decided to create a new alpine skiing dataset featuring semi-professional athletes for which videos are publicly available. 

\begin{figure}[t]
		\begin{center}
			\includegraphics[width=0.95\linewidth]{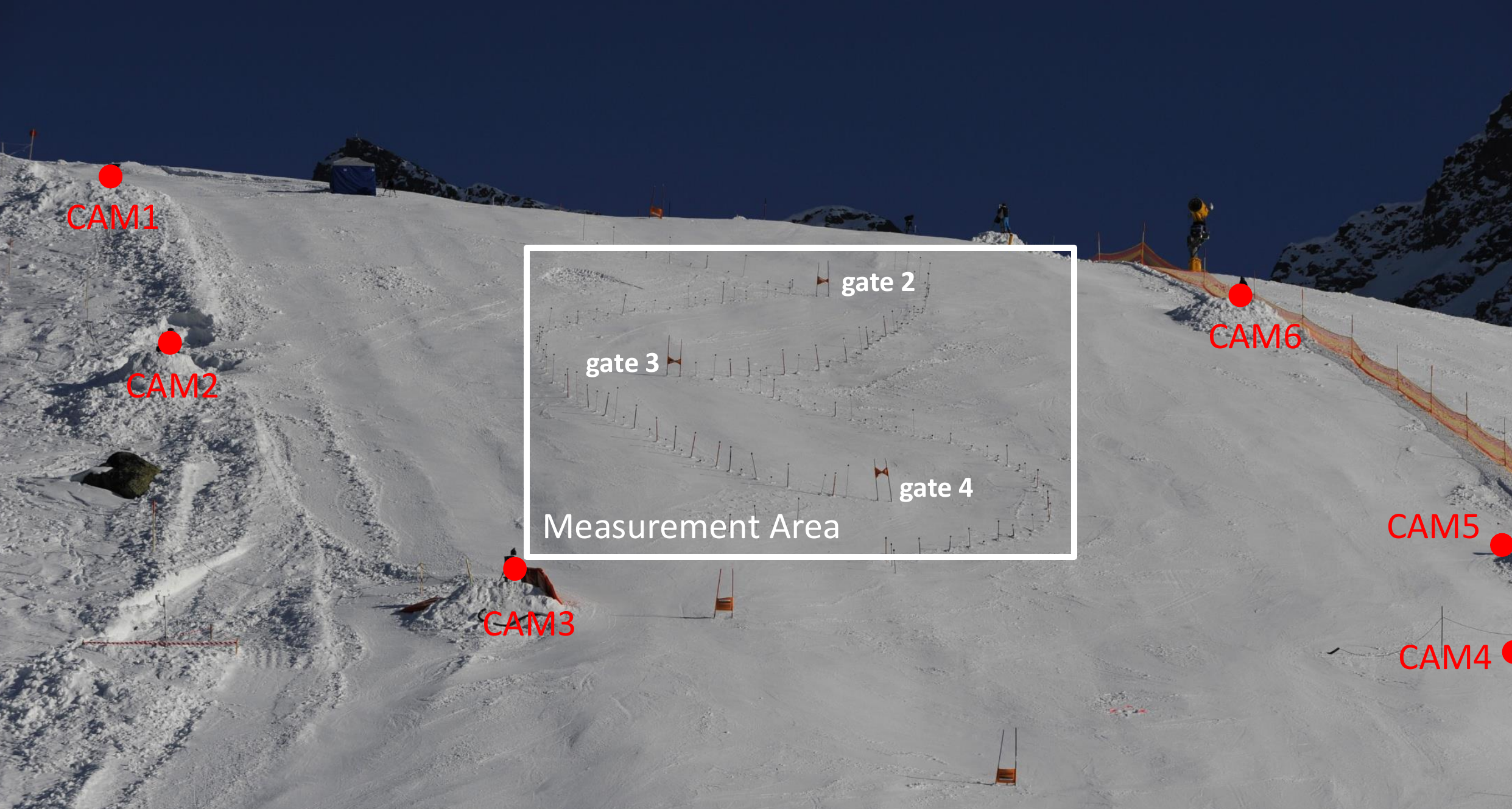}
		\end{center}
		\vspace*{-0.5cm}
		\caption{Setup overview of the multi-view skiing dataset.}
		\label{fig:kuetai_setup}
\end{figure}

For the purpose of evaluating 3D pose estimation methods and comparing to related work, we used a manually annotated multi-view (\emph{MV-Ski}) pan-tilt-zoom alpine skiing dataset~\cite{SpoerriInjuryPrevention, AlpineSkiingKinematics, RhodinUnsupervisedSki}. It features 6 professional athletes on a Giant Slalom slope with three turns, filmed by six cameras that are arranged in a circle around the center of the track as shown in Figure~\ref{fig:kuetai_setup}. 2D joint locations were manually annotated. Calibration points around the track served to calculate the camera parameters, specifically the intrinsic and extrinsic camera matrices. From this, global ground truth 3D poses were triangulated.

While the \emph{MV-Ski} dataset is well suited for developing semi-supervised models~\cite{RhodinUnsupervisedSki}, the fact that it only features 6 athletes in similar suits performing the run on the same slope with the same camera angles makes methods trained on it unable to generalize to different skiing settings.

\section{New single-view alpine skiing dataset}\label{sec:new_dataset}
\begin{figure}[t]
		\centering
		\begin{subfigure}[b]{0.23\textwidth}
			\centering
			\includegraphics[width=\textwidth]{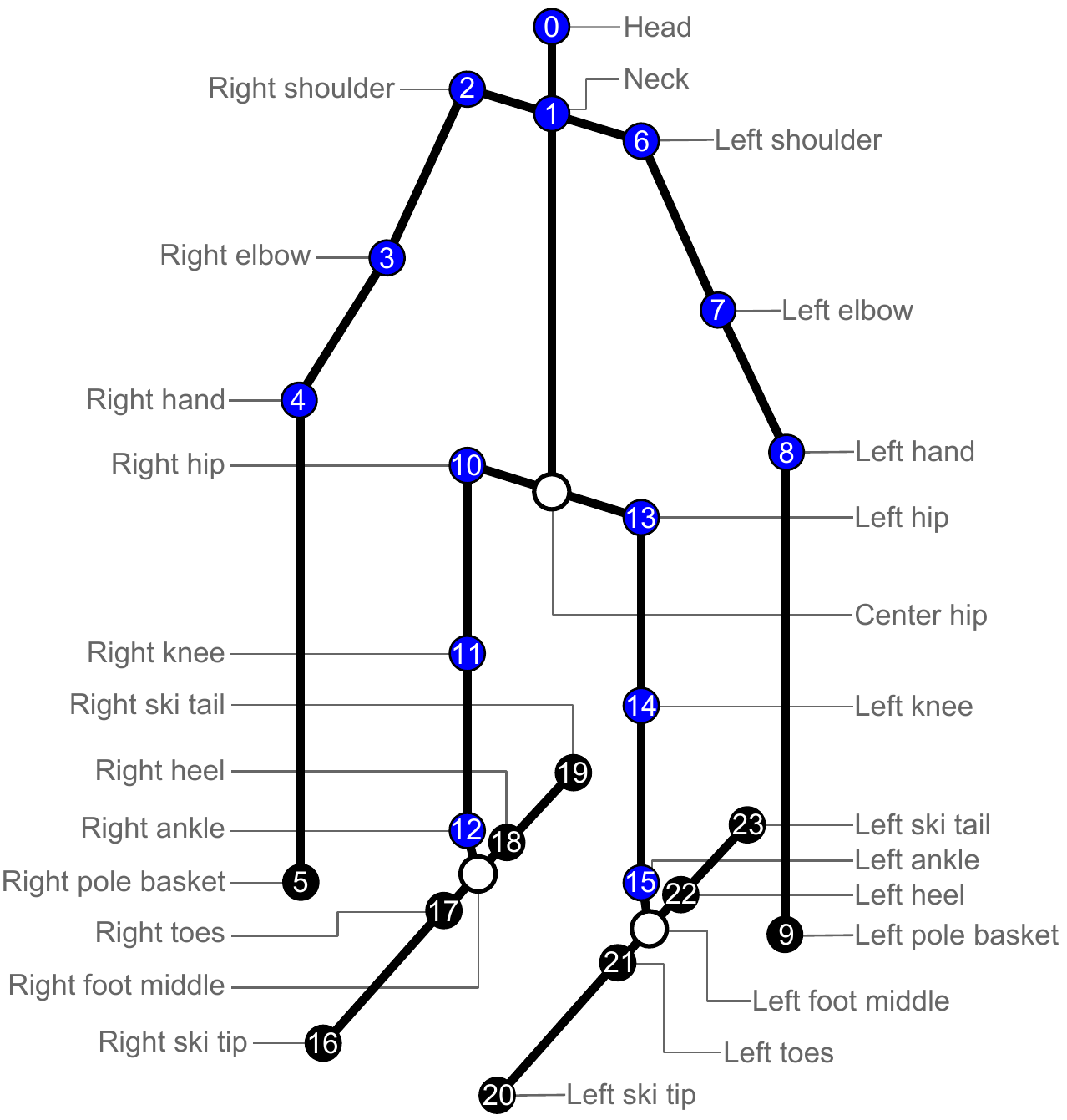}
			\caption{Skier joints}
			\label{fig:skier_joints}
		\end{subfigure}%
		\begin{subfigure}[b]{0.237\textwidth}
			\centering
			\includegraphics[width=\textwidth]{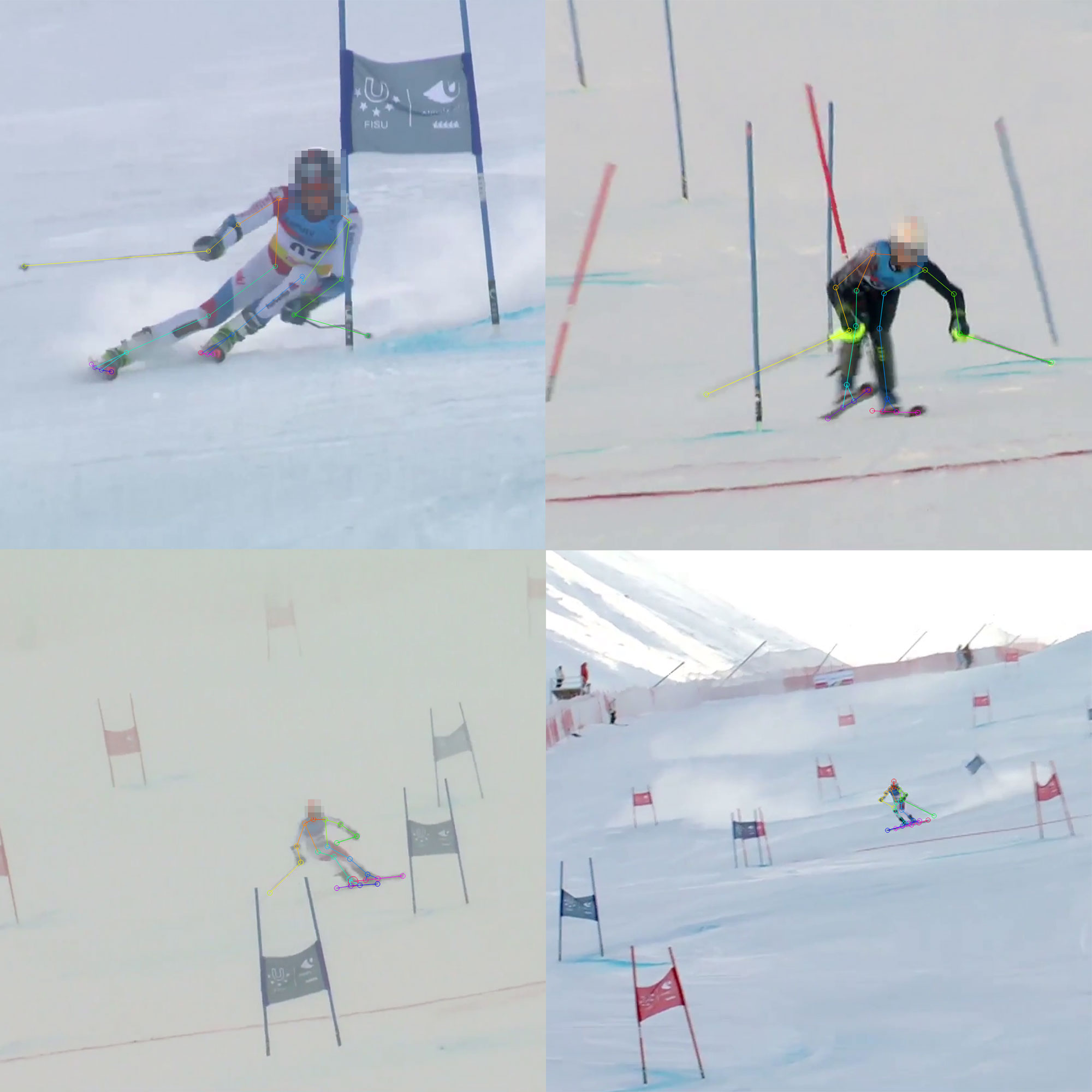}
			\caption{Example images}
			\label{fig:alpine_examples}
		\end{subfigure}
		\vspace*{-0.2cm}
		\caption[Caption]{Figure~\subref{fig:skier_joints} shows the 24 annotated joints, with the subset of body joints marked in blue. The white circles are non-annotated helper joints. Figure~\subref{fig:alpine_examples} shows four annotated images of the new dataset featuring various conditions.}
	\end{figure}

	To facilitate generalization of marker-less ski motion capture to new environments, we create a large single view (\emph{SV-Ski}) dataset for alpine skiing. We downloaded 16 alpine skiing videos that were posted on Youtube under the Creative Commons license, featuring mainly semi-professional ski racers from many different perspectives in various weather conditions. Those videos were split into 147 training and 11 validation sequences of various lengths, from which frames were sampled in fixed intervals ranging from 0.3 to 10 seconds, depending on the discipline. In total, 1982 images were sampled and annotated with 24 2D key points, as depicted in Figure~\ref{fig:skier_joints}, of which 1830 were used as training and 152 as validation images. The dataset comprises at least 32 unique athletes in 5 unique locations and various conditions (see Figure~\ref{fig:alpine_examples}) and is made available online \footnote{Single view alpine skiing dataset: \url{https://cvlab.epfl.ch/ski-2dpose-dataset/}} for further research.
	
\begin{figure}[t]
		\begin{center}
			\includegraphics[width=0.95\linewidth]{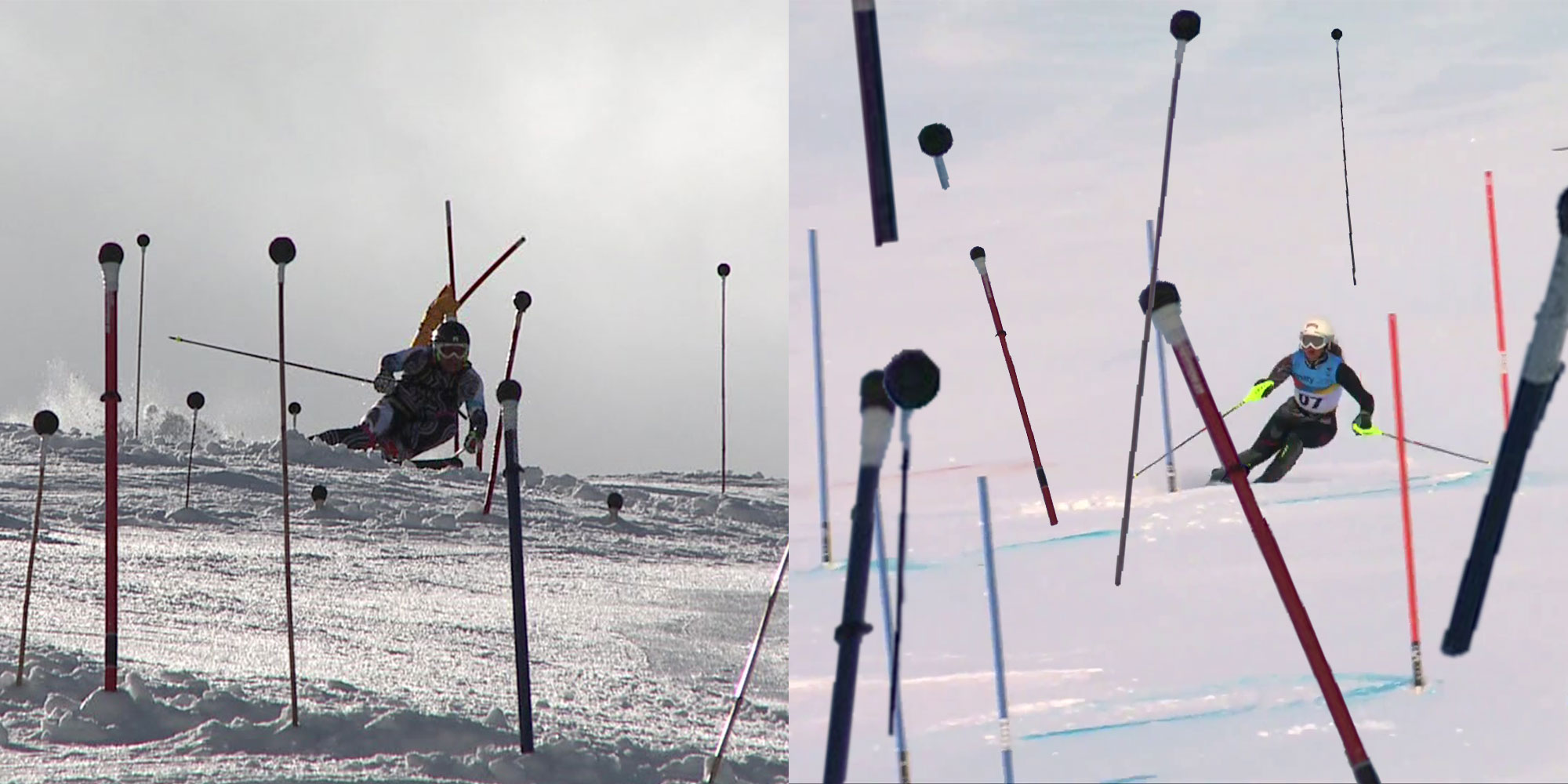}
		\end{center}
		\vspace*{-0.5cm}
		\caption{Left: Example image showcasing calibration poles. Right: Augmented alpine skiing image.}
		\label{fig:kuetai_augmented_examples}
	\end{figure}

	\paragraph{Calibration pole augmentation.}
	As this newly created Alpine dataset does not feature any calibration poles like the \emph{MV-Ski} dataset, evaluating a 2D pose estimation algorithm that was only trained on this will produce significant outliers, particularly for the ski poles. One way to improve robustness on the \emph{MV-Ski} dataset is to augment SV-training images with randomly superimposed cutouts of various calibration poles, see Figure~\ref{fig:kuetai_augmented_examples}. At training time, we uniformly sample $\mathcal{U}(0,20)$ randomly selected poles and place them uniformly over the image. The poles are scaled by $\mathcal{U}(0.5,2.5)$ and rotated $\mathcal{U}(-15,15)$ degrees. We compare this method to adding one \emph{MV-Ski} sequence to the training of OpenPose.

\section{Method}
	\begin{figure*}[t]
		\begin{center}
			\includegraphics[width=0.92\linewidth]{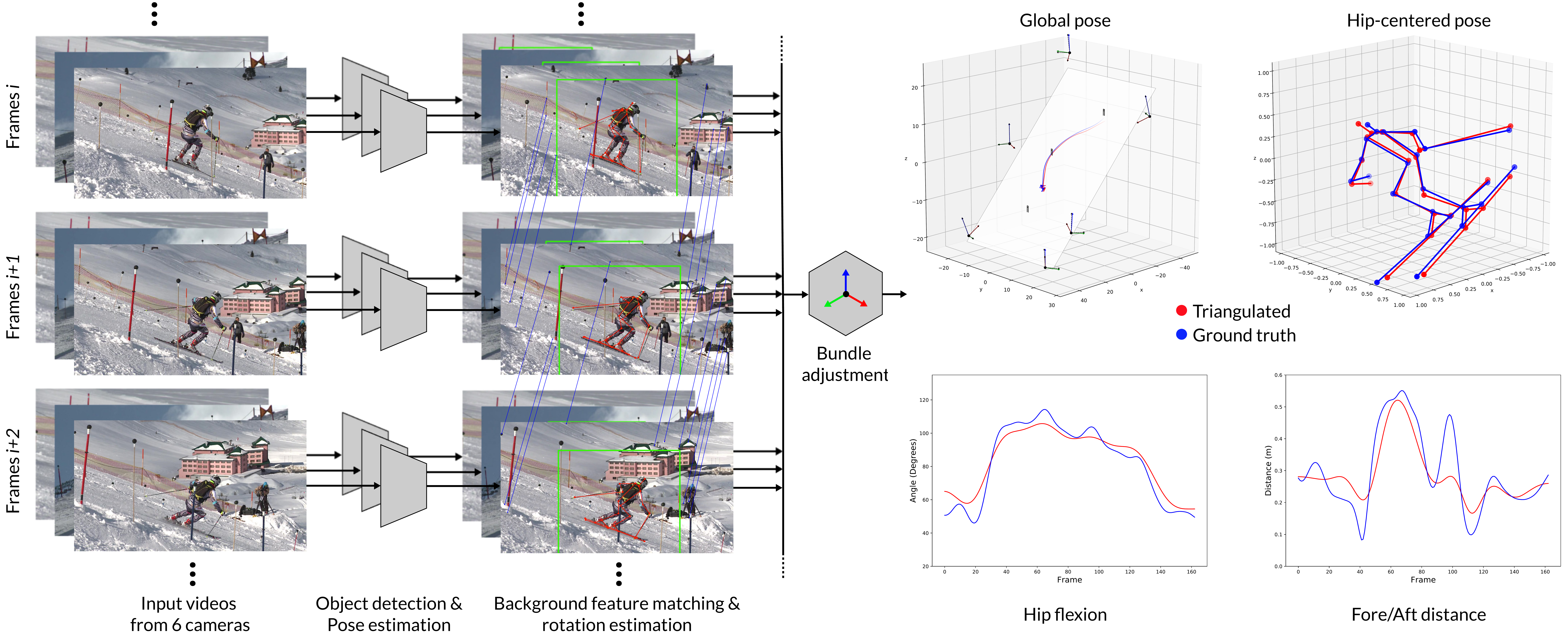}
		\end{center}
		\vspace*{-0.5cm}
		\caption{Method overview -- Images from up to 6 different cameras are preprocessed to find 2D joint locations by first cropping a bounding box around the athlete and then running a pose estimation network on it. 2D poses from all cameras and all frames are used in a bundle adjustment approach to triangulate the athlete's global 3D pose. Additionally, camera rotations can be estimated by tracking background features over consecutive frames. Coaches can extract useful information about biomechanical variables like the hip flexion or fore/aft distance along the athlete's trajectory from the 3D pose.}
		\label{fig:pipeline}
	\end{figure*}

	Our goal is to take as input a set of synchronized video streams of the same athlete filmed from different angles and estimate the global and articulated 3D pose. We target an easy-to-setup solution: The cameras are assumed to be intrinsically calibrated, have known relative position, but unknown orientation. 
	The advantage is that this setup allows us to use consumer cameras without specialized hardware for angular readout --- handheld recording is conceivable. The position requirement might sound restrictive, however, when using up to three cameras a simple distance measure between pairs of cameras is sufficient to determine their relative position. Furthermore, the intrinsics for fixed-focal length cameras only have to be calibrated once. Still, professional PTZ-cameras that have been calibrated for different zoom levels can be used as well.
	
	We develop our approach for estimating the gloabal articulated 3D pose in two steps. In the first, we assume the rotation matrices as known, while in the second we jointly optimize for 3D pose and camera rotations. To go from images to 3D pose, we propose a multi-staged approach as shown in Figure~\ref{fig:pipeline}, where 2D pose detections are generated from cropped images around the athlete and then 3D poses are optimized to best fit all localized 2D joints. Generating 2D estimations first allows us to analyze potential detection weaknesses when using a new dataset, before developing a method for 3D joint detection.

	First, we train and run an object detection network \cite{SSD} on each video stream to generate a tight square bounding box around the primary athlete, which effectively excludes persons in the background. Outliers are filtered out and bounding box detections are temporally smoothed. The 2D pose estimation network \cite{OpenPose} is subsequently trained and run on the square crop, generating joint heatmaps, from which 2D joint key points are extracted. The 2D detections from all cameras are then incorporated into a bundle adjustment method which reconstructs the underlying 3D pose of the skier. This optimization includes our core contribution on conditioning camera motion on tracked features.

\subsection{Athlete detection}

Sports videos are often captured from a distance and contain sequences of very different zoom levels, resulting in the athlete size in the image ranging from frame-filling to only making up a very small portion of the frame. While pose estimation networks like OpenPose run images through their network at multiple scales to account for this fact, athletes can be so small in images that detection fails completely. Even when the skier fills a good portion of the frame and the input scale is right for the pose estimation network, other people on the slopes or other high-contrast objects in the background can lead to wrong detections when only interested in the pose of the main subject. We therefore first detect a tight bounding box around the athlete, resulting in the pose estimation network always receiving examples of the same scale.

In general, object detection is the process of localizing occurrences of certain classes in images and drawing a tight bounding box around them. In recent years, several Deep Learning approaches made great advances in terms of accuracy and detection speed. Liu \etal propose the Single Shot MultiBox Detector (SSD)~\cite{SSD} generating scores for the presence of an object in predefined bins and then adjusting the bins to better match the object shape. Detections from multiple feature maps and different resolutions are then combined to allow for detection of various sizes in one single network stage. For this task we chose the SSD~\cite{SSD} network for its good performance and low computational overhead during both training and test time.

As the SSD network is a single-image detector and does not incorporate temporal information from the fact that we are dealing with videos, detections between frames can suffer from jitter and outliers if other people are present in the scene. To remove strong outliers, we replace all detections whose center deviates beyond the bounding box of the last correctly detected frame by the latter. Jitter and varying sizes of the bounding boxes are dealt with by applying temporal Gaussian smoothing to the center location of the crops and their respective side lengths with parameters $\sigma_{center} = 10$ and $\sigma_{side} = 50$ frames. Finally, side lengths are scaled by a factor of 1.5 for the newly created \emph{SV-Ski} and 2 for the \emph{MV-Ski} dataset and all crops get resized to 500 x 500 pixels. This was done because the SSD network tended to produce too small bounding boxes on the unseen \emph{MV-Ski} dataset. Training specifications for the SSD network are described in the supplementary material.

\subsection{2D pose estimation}
	Given an input image $\bm{I} \in  \mathbb{R}^{w \times h \times 3}$ of width $w$ and height $h$, the task of 2D pose estimation is to compute x and y coordinates for every joint $j \in \{1,...,N_J\}$. The OpenPose network by Cao \etal~\cite{OpenPose} returns for every joint $j$ a confidence/heat map $\bm{C}_j \in \mathbb{R}^{w \times h}$ and for each limb/bone $l \in \{1,...,N_L\}$ a Part Affinity Field (PAF) $\bm{B}_l \in \mathbb{R}^{w \times h \times 2}$, with every point in $\bm{B}_l$ encoding a vector describing limb orientations. Using the PAF as an indicator for which joints in the heatmaps belong together, the poses of multiple people can be efficiently differentiated. We finetune this model for the extended set of ski pose joints, including ski tip and tail, and pole positions. Because in our problem we are focusing solely on the pose estimation of a single athlete on the slope and not multiple people, we don't rely on some of the multi-person detection advantages that PAF's bring to the table. Indeed, for this task we take the maximum location $\bm{p}_j^* \in \mathbb{R}^2$ as $\bm{p}_j^* = \argmax_{w,h} \bm{C}_j$ of each confidence map for each joint~$j$.
	Training specifications for the OpenPose network are described in the supplementary material.

\para{Refining SSD bounding boxes using OpenPose.}
	Because of the Gaussian smoothing of detected SSD bounding boxes, drifts in the crop with respect to the athlete's center may still be present. Athletes often extend their arms and poles outwards for balance in difficult terrain, which may cause the thin ski poles to exit the bounding box when drifts in the smoothed crops are present.
	To remedy this, we run OpenPose on the generated crops and take the median of the computed joint positions as the new center for each frame. Then, we apply a weaker Gaussian smoothing pass to the center locations and side lengths with parameters $\sigma_{center} = 5$ and $\sigma_{side} = 5$, and re-run OpenPose.

\subsection{3D pose estimation}
	The last step in our approach is estimating the 3D poses of the skier using a bundle adjustment optimization method with the detected 2D joint locations. We take the OpenPose output $\bm{p}_j^{f,c} \in \mathbb{R}^2$ from all cameras $c \in \{1,...,N_C\}$, over all frames $f \in \{1,...,N_F\}$, for each joint $j \in \{1,...,N_J\}$ and reconstruct the respective underlying 3D joint positions $\bm{P}_j^f \in \mathbb{R}^3$ in global space. Let us denote the complete 3D pose at time $f$ as $\bm{P}^f \in \mathbb{R}^{N_J \times 3}$.

\para{Cosine basis parametrization.}
    One way to go about the bundle adjustment would be to directly optimize for the points $\bm{P}$ in 3D space. We would jointly try to optimize $f \times N_J \times 3$ unrestricted parameters, meaning poses in neighboring frames are not guaranteed to transition in a smooth motion. This is usually countered by penalizing accelerations measured by finite differences.

    Instead of letting our 3D points be completely free and add a smoothness term after the fact, we chose to describe them using a parametrization that is inherently smooth over time \cite{akhter2012bilinear}. We model the motion using the inverse discrete cosine transform (IDCT), meaning that it is the result of a sum of $N_{\Pi}$ cosine waves, scaled by coefficients $\bm{\Pi}_{j,d} \in \mathbb{R}^{N_{\Pi}}$ for each joint $j \in \{1,...,N_J\}$ and its dimension $d \in \{x,y,z\}$. The 3D pose can then be reconstructed using
    \begin{equation}\label{eq:idct}
        \bm{P}_{j,d}^{f} = \frac{\bm{\Pi}_{j,d}^{0}}{2} \sum_{n=1}^{N_{\Pi} - 1} \bm{\Pi}_{j,d}^{n} \cos \left[ \frac{\pi n}{N_{\Pi}} \left( f + \frac{1}{2} \right) \right] \;, 
    \end{equation}
    for $f \in \{0,...,N_F - 1\}$.
    
    The lower we set $N_{\Pi}$, the smoother the motion will be, as we are only using low-frequency cosine waves, but we might not be able to reproduce actual fast changes in movement. On the other hand, if we increase the number of coefficients, we could approximate more complex motions but risk picking up high-frequency noise. With known rotation matrices, we set $N_{\Pi}$ to 25, while with uncalibrated cameras we set it to 11. These values were chosen empirically for a good balance between smoothness and accuracy.

\subsection{Using known camera rotations}
	In this section, we assume that the parameters for all frames $f$ and all cameras $c$ are known. Specifically, this means we know the intrinsic matrix $\bm{K}^{f,c} \in \mathbb{R}^{3 \times 3}$, the matrix describing world to camera rotation $\bm{R}^{f,c} \in \mathbb{R}^{3 \times 3}$ and camera location $\bm{t}^{f,c} \in \mathbb{R}^{3}$. Using the extrinsics ${[\bm{R}^{f,c}\mid\bm{t}^{f,c}]}$, the transformation of a world coordinate point $\bm{P}_j^{f,w}$ to camera $c$'s coordinate frame is given by
	\begin{equation}
	\bm{P}_j^{f,c} = \bm{R}^{f,c} \bm{P}_j^{f,w} + \bm{t}^{f,c} \;.
	\end{equation}
	The projection $\hat{\bm{p}}_j^{f,c} \in \mathbb{R}^3$ (in homogeneous coordinates) of point $\bm{P}_j^{f,c}$ onto camera $c$'s image plane is then given by
	\begin{equation}
	\hat{\bm{p}}_j^{f,c} = \bm{K}^{f,c} \bm{P}_j^{f,w} \;.
	\end{equation}
	The homogeneous point $\hat{\bm{p}}_j^{f,c}$ can then be transformed to the Euclidean point $\tilde{\bm{p}}_j^{f,c} \in \mathbb{R}^2$ by dividing by the last coordinate. Finally, denote the complete projection from world coordinates to an image plane as
	\begin{equation}
	\pi_c(\bm{P}_j^{f,w}) = \tilde{\bm{p}}_j^{f,c} \;.
	\end{equation}

	3D reconstruction is done using a bundle adjustment approach, where we optimize an energy function
	\begin{equation}
	\argmin_{\bm{\Pi}} E(\bm{\Pi}, \bm{K}, \bm{R}, \bm{t}) \;,
	\end{equation}
	that includes a reprojection error, as well as priors on the human body defined as
	\begin{equation}
	E(\bm{\Pi}, \bm{K}, \bm{R}, \bm{t}) = \lambda_{rep} E_{rep} 
		+ \lambda_{limbs} E_{limbs} \;.
	\end{equation}

	\para{Reprojection term.}
	The 3D joint location estimations are iteratively updated by gradient descent such that when projected to each camera plane, they are as close as possible to the 2D joint locations. If we had perfectly consistent 2D localizations, a simple least-squares bundle adjustment process with decent initializations would yield very good results. In our case, 2D detections sometimes contain high per-joint pixel errors, and we use a robust norm that also incorporates the detection confidence, similar to Takahashi \etal~\cite{Takahashi}. We write the reprojection energy term as
	\begin{equation}
	\scalemath{0.83}{
	E_{rep}(\bm{\Pi}, \bm{K}, \bm{R}, \bm{t}) = \frac{1}{N_F N_C N_J} \sum_{f=1}^{N_F} \sum_{c=1}^{N_C} \sum_{j=1}^{N_J} g(\pi_c (\bm{P}_j^{f,w}), \bm{p}_j^{f,c})
	},
	\end{equation}
	with $\bm{P}_j^{w} = \mathrm{IDCT}(\bm{\Pi}_{j})$ in DCT encoding.
	The distance
	\begin{equation}
	g(x,y) = (n(0) - n(e_{rep}(x,y))) e_{rep}(x,y)
	\end{equation}
	re-weights the scaled reprojection errors
	\begin{equation}
	e_{rep}(x,y) = ||(x-y) \bm{C}(y)||_2 \;,
	\end{equation}
	where $n(x)$ denotes the normal distribution's probability density function $N(0, \sigma^2)$ and $\bm{C}(y)$ the heatmap probability value at point y. Using this norm with $\sigma^2=100$, outlier points have negligible influence on the energy function.

	\para{Human prior term.}
	We would like all limbs $(i,j) \in Limbs$ to consistently have the same lengths $\ell(i,j)$ over time. To this end, we minimize the difference between the estimated and the known limb lengths,
	\begin{equation}
	\scalemath{0.82}{
	E_{limbs}(\bm{\Pi})= \frac{1}{N_F} \sum_{f=1\dots N_F} \sum_{i,j} \left(\| \bm{P}_i^{f,w} - \bm{P}_j^{f,w} \|_2 - \ell(i,j)\right)^2
	},
	\end{equation}
	with $\bm{P}_j^{w} = \mathrm{IDCT}(\bm{\Pi}_{j})$ in DCT encoding. The limb lengths $\ell(i,j)$ were taken from the ground truth data, but can also be measured manually on the athletes.

	\para{Optimization and parameters.}
	When optimizing for absolute 3D positions, all points were initialized in the center between all cameras, with an additional random spread of $\mathcal{U}(-10,10)$ meters. We used the L-BFGS~\cite{L-BFGS}, a quasi-Newton optimization algorithm,
	with step length 0.05, running it for 100 outer iterations, with at most 20 inner iterations per optimization step. The energy terms were scaled by $\lambda_{rep} = 80$ and $\lambda_{limbs} = 1$.

\subsection{Estimating camera rotations}
	In the same way we optimized the 3D pose positions, it is possible to freely optimize other parameters such as the camera's rotation~\cite{Takahashi}. 
	We again use the IDCT as in Equation~\ref{eq:idct} to compactly describe the Euler angles of the camera rotations using $N_{\Gamma}=11$ coefficients $\bm{\Gamma}$. The objective then becomes
	\begin{equation}
	\argmin_{\bm{\Pi}, \bm{\Gamma}} E(\bm{\Pi}, \bm{K}, \bm{\Gamma}, \bm{t}) \;.
	\end{equation}
	
	\para{Initialization.}
	Like with the 3D pose, the camera angles can be parametrized by a low-dimensional cosine basis and iteratively updated to the correct ones by gradient descent. A problem with this approach is however initialization. If the cameras face in randomly initialized directions, it is unlikely that the optimization objective can converge to a desirable solution. We instead propose a bootstrapping step, where in every gradient descent iteration only the 3D poses are optimized, while the cameras are adjusted to always point to the center of the estimated poses.

	More specifically, in the beginning we initialize all 3D pose positions around the center of all cameras with a random spread of $\mathcal{U}(-1,1)$ meters to be largely independent of any specific sport. For every camera we then compute the look-at rotation matrix, with the target being the mean location of the 3D pose and the camera's up direction being the global z-axis. A problem with the look-at matrix is, that the person is originally not necessarily in the middle of the image. To solve this, we first compute the horizontal and vertical relative position of the skier in the 2D image. From the camera intrinsics, we know the Field of View (FoV) and can then pan and tilt the rotation matrix in the opposite direction of the calculated horizontal and vertical FoV shift.

	When optimizing with this method for 25 outer iterations, we get a very rough estimate for the real 3D pose positions and camera rotation matrices, which serves as an initialization for joint optimization of all parameters. 
	
	\para{Homography camera rotation differences.}
	From this initial estimate, we could potentially optimize for 3D poses and camera rotations jointly, but preliminary tests (Table~\ref{tab:comparison}) have shown that doing this with only the 2D pose estimates as information for the optimization does not yield accurate results. What we propose instead is to use the background information and the fact that we are dealing with PTZ-cameras to our advantage. Since PTZ-cameras are fixed in space and can therefore only rotate, detected 2D points, say of consecutive frames $f$ and $f+1$ in camera $c$, are related~\cite{MultipleViewGeom} by a homography matrix $\bm{H}^{f,c}$ satisfying 
	\begin{equation}
	    \bm{H}^{f,c} = \bm{K}^{f+1,c} \Delta \bm{R}^{f,c} (\bm{K}^{f,c})^{-1} \;,
	\end{equation}
	with $\Delta \bm{R}^{f,c}$ being the relative rotation between the images and $\bm{K}^{f,c}$ and $\bm{K}^{f+1,c}$ the intrinsics of the first and second frame, respectively.
	
\begin{figure}[t]
		\begin{center}
			\includegraphics[width=\linewidth]{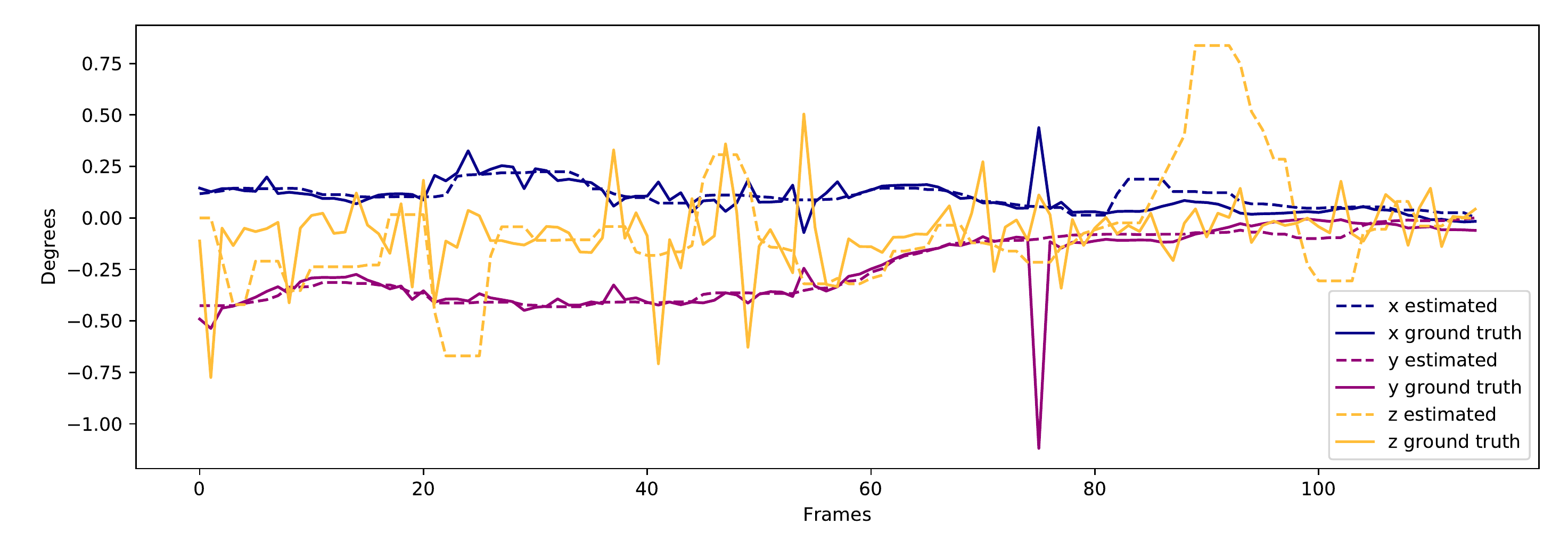}
		\end{center}
		\vspace*{-0.5cm}
		\caption{Estimated change in rotation $\Delta \bm{R}$ for one camera.}
		\label{fig:rotation_est}
    \end{figure}

	Because we assume the intrinsics as known, to find the rotation we only need to compute the homography. For this, we collect features in consecutive images using an ORB detector~\cite{ORB, CVORBDetector} and match them by minimizing the Hamming norm~\cite{CVHamming}. We only consider those points lying outside the bounding box predicted by the SSD network, to exclusively track features in the static background. The corresponding points are then used to find the homography matrix~\cite{CVFindHomography}, excluding outliers with RANSAC. Finally we compute the rotation between any consecutive frames $f$ and $f+1$
	\begin{equation}
	    \Delta \bm{R}^{f,c} = (\bm{K}^{f+1,c})^{-1} \bm{H}^{f,c} \bm{K}^{f,c} \;.
	\end{equation}
	For cameras with a fixed position, $\Delta \bm{R}^{f,c}$ is a pure rotation matrix. In practice, slight camera movements and noisy inputs to the homography computation will not produce perfect rotation matrices, but we found that extracted Euler angles are not far off their ground truth, as shown for one example sequence in Figure~\ref{fig:rotation_est}. Notice that since the camera has a large focal length, pitch and yaw changes in the x and y axes are much more significant than rolling in the z axis, and are also estimated more accurately. Outliers in the rotation differences are removed by a median filter with size 7 and smoothed using a Gaussian with standard deviation 3.
	
	\para{Bundle adjustment with rotation differences.}
	In our energy function, we add a term
	\begin{equation} \label{eq:error_rot}
	\scalemath{0.82}{
    E_{rot}(\bm{\Gamma}) = \frac{1}{N_F N_C} \sum_{f=1}^{N_F-1} \sum_{c=1}^{N_C} \norm{\Delta \bm{R}^{f,c} - \left( \bm{R}^{f+1,c} \left( \bm{R}^{f,c} \right)^{\intercal} \right) }_2
    },
    \end{equation}%
    with $\bm{R}^{c} = \mathrm{IDCT}(\bm{\Gamma}^{c})$ in DCT encoding, minimizing the norm between the measured $\Delta \bm{R}^{f,c}$ and estimated consecutive camera rotations $\bm{R}^{f+1,c} \left( \bm{R}^{f,c} \right)^{\intercal}$.
    By minimizing this term, we enforce that the estimated relative camera motion matches the one measured by optical flow, which enables the bundle adjustment to find the absolute rotation more easily.
    Note that this formulation is robust to measurement noise and the estimated homographies not being pure rotations, since the Euler-angle representation ensures that $\bm{R}$ is a proper rotation matrix.
	We further optimize for $\bm{\Pi}$ and $\bm{\Gamma}$ over 1500 outer iterations. The energy function was weighted with $\lambda_{rep} = 500$, $\lambda_{limbs} = 1$ and $\lambda_{rot} = 10000$.

\section{Results}
	In this section, we report the performance of our motion capture algorithm on the task of ski performance analysis. First, we quantify the accuracy gain brought about by the new \emph{SV-Ski} dataset, optical-flow guided rotation estimation, and other model choices. Second, we compare against existing monocular methods for 3D human pose estimation on the public \emph{MV-Ski} dataset.
	
	\para{Metrics.}
	We report the widely-used Percentage of Correct Key points (PCK) metric, the fraction of predicted joint positions that is within one head-neck distance to the ground truth, as well as the Mean Per Joint Position Errors (MPJPE), the mean euclidean distance to the ground truth. For 2D keypoints, MPJPE is measured in normalized image coordinates ranging from 0 to 1, for 3D in meters (m).

	In addition we, analyze the mean absolute error (MAE) of skiing-specific metrics that are widely used in performance analysis~\cite{SpoerriInjuryPrevention, AlpineSkiingKinematics}, such as the center of mass (CoM), knee angle, and lean angle, where the latter is measured in the plane orthogonal to the skiing direction. These are defined formally in the supplementary material.
	For the MPJPE and CoM, we analyze the \emph{Global} error computed in world coordinates, the \emph{Centered} error measured relative to the hip, and in \emph{Normalized} coordinates where the scale of the prediction is adjusted to the ground truth in the least squares sense before error computation. Monocular methods can only estimate the latter, as scale and depth is ambiguous without knowing the athletes height.
	All metrics are computed over the available test sequences and the mean and standard deviation (std) across all frames is reported.
	
	\para{Test sets.}
	\emph{MV-Ski-test} contains two runs of a skier not contained in the training set, totalling to 1674 frames. \emph{SV-Ski-test} comprises 152 images that are strictly excluded from training.

\subsection{2D pose estimation}\label{sec:2D_results}
	We trained OpenPose using four different dataset configurations. First we only trained it on the newly created SV-Ski dataset, which we then augmented with calibration poles. We then initialized the network using pretrained weights from the MPII Human Pose dataset, and finally added one \emph{MV-Ski} sequence using four camera angles.

	\begin{table}[t]
    	\begin{subtable}[h]{0.49\columnwidth}
    		\centering
    		\resizebox{0.95\columnwidth}{!}{%
    		\begin{tabular}{|l|c|c|}
    		    \multicolumn{3}{c}{{\bf(a)} SV-Ski training}\\
    			\hline
    			Test dataset & PCK & MPJPE $\pm$ std\\
    			\hline\hline
    			SV-Ski all & 92.69 & 0.0195 $\pm$ 0.0638 \\
    			SV-Ski body & 94.88 & 0.0132 $\pm$ 0.0416 \\
    			MV-Ski all & 51.37 & 0.1064 $\pm$ 0.1297 \\
    			MV-Ski body & 58.26 & 0.0835 $\pm$ 0.1092 \\
    			\hline
    		\end{tabular}
    		}
    	\end{subtable}
    	\hfill
    	\begin{subtable}[h]{0.49\columnwidth}
    		\centering
    		\resizebox{0.95\columnwidth}{!}{%
    		\begin{tabular}{|l|c|c|}
    		    \multicolumn{3}{c}{{\bf(b)} SV-Ski Augmented training}\\
    			\hline
    			Test dataset & PCK & MPJPE $\pm$ std\\
    			\hline\hline
    			SV-Ski all & 95.77 & 0.0807 $\pm$ 0.0488 \\
    			SV-Ski body & 97.91 & 0.0625 $\pm$ 0.0346 \\
    			MV-Ski all & 65.51 & 0.0137 $\pm$ 0.1236 \\
    			MV-Ski body & 73.01 & 0.0092 $\pm$ 0.1074 \\
    			\hline
    		\end{tabular}
    		}
    	\end{subtable}
    	\smallskip\vspace{0.05cm}
    	\begin{subtable}[h]{0.49\columnwidth}
    		\centering
    		\resizebox{0.95\columnwidth}{!}{%
    		\begin{tabular}{|l|c|c|}
    		    \multicolumn{3}{c}{{\bf(c)} SV-Ski Aug. + weight init. MPII}\\
    			\hline
    			Test dataset & PCK & MPJPE $\pm$ std\\
    			\hline\hline
    			SV-Ski all & \bf 96.76 & \bf 0.0119 $\pm$ 0.1268 \\
    			SV-Ski body & \bf 98.36 & \bf 0.0081 $\pm$ 0.1077 \\
    			MV-Ski all & 70.10 & 0.0755 $\pm$ 0.0429 \\
    			MV-Ski body & 76.83 & 0.0577 $\pm$ 0.0268 \\
    			\hline
    		\end{tabular}
    		}
    		\label{tab:2D_results_c}
    	\end{subtable}
    	\hfill
    	\begin{subtable}[h]{0.49\columnwidth}
    		\centering
    		\resizebox{0.95\columnwidth}{!}{%
    		\begin{tabular}{|l|c|c|}
    		    \multicolumn{3}{c}{{\bf(d)} SV-Ski Aug. + init. MPII + MV-Ski}\\
    			\hline
    			Test dataset & PCK & MPJPE $\pm$ std\\
    			\hline\hline
    			SV-Ski all & 96.51 & 0.0119 $\pm$ 0.0431 \\
    			SV-Ski body & 97.81 & 0.0087 $\pm$ 0.0296 \\
    			MV-Ski all & \bf 78.11 & \bf 0.0627 $\pm$ 0.1275 \\
    			MV-Ski body & \bf 83.12 & \bf 0.0507 $\pm$ 0.1133 \\
    			\hline
    		\end{tabular}
    		}
    	\end{subtable}
    
    	\caption{{\bf 2D pose estimation results} on \emph{SV-Ski} and \emph{MV-Ski} with four different dataset configurations used for training.}
    	\label{tab:2D_results}
    \end{table}

    In Table~\ref{tab:2D_results} we analyze different training and test splits, using \emph{all} keypoints and also just the \emph{body} joints, the 14 joints 0-4, 6-8, and 10-15 shown in Figure~\ref{fig:skier_joints}. In the case of the \emph{SV-Ski-test} set we have information about joint visibility and invisible joints were not counted in the PCK results. The data augmentation and taking one \emph{MV-Ski} sequence for training brought the biggest gain in accuracy. The accuracy on the \emph{SV-Ski-test} set improved overall, but not as much because it was already very high. See the supplementary material for the same comparisons, displayed graphically.

\subsection{3D pose estimation}
    \begin{table*}[!htbp]%
    	\begin{center}%
    	\resizebox{.89\linewidth}{!}{%
    		\begin{tabular}{|l|c|c|c|c|c|c|}%
    			\hline
    			Metric & Ours-calibrated & Ours-uncalibrated & Ours-A (calibrated) & Ours-B (uncalibrated) & C & D \cite{RhodinUnsupervisedSki}\\
    			\hline\hline
    			Global MPJPE [m]          &\bf0.092 $\pm$ 0.091  &   0.701 $\pm$ 0.219  &   0.096 $\pm$ 0.120 &  7.590 $\pm$ 4.946  &  n/a              &  n/a   \\
    			Global Body MPJPE [m]     &   0.060 $\pm$ 0.046  &   0.688 $\pm$ 0.201  &\bf0.056 $\pm$ 0.033 &  7.490 $\pm$ 4.792  &  n/a              &  n/a   \\
    			Centered MPJPE [m]        &\bf0.077 $\pm$ 0.087  &   0.090 $\pm$ 0.085  &   0.087 $\pm$ 0.122 &  0.459 $\pm$ 0.960  &  n/a              &  n/a   \\
    			Centered Body MPJPE [m]   &\bf0.045 $\pm$ 0.030  &   0.071 $\pm$ 0.053  &   0.050 $\pm$ 0.034 &  0.355 $\pm$ 0.383  &  n/a              &  n/a   \\
    			Normalized MPJPE [m]      &   0.075 $\pm$ 0.083  &   0.087 $\pm$ 0.082  &   0.087 $\pm$ 0.117 &  0.232 $\pm$ 0.238  &\bf0.07            &  n/a   \\
    			Normalized Body MPJPE [m] &\bf0.039 $\pm$ 0.025  &   0.051 $\pm$ 0.035  &   0.042 $\pm$ 0.029 &  0.132 $\pm$ 0.103  &  n/a              &  0.081 \\
    			Global CoM Error [m]      &   0.05  $\pm$ 0.04   &   0.78  $\pm$ 0.24   &\bf0.05  $\pm$ 0.02  &  8.93  $\pm$ 5.40   &  n/a              &  n/a   \\
    			Global speed MAE [m/s]    &\bf0.74  $\pm$ 1.76   &   1.87. $\pm$ 2.64   &   0.45  $\pm$ 1.08  & 31.35  $\pm$22.45   &  n/a              &  n/a   \\
    			Knee flexion MAE [deg]    &   3.96  $\pm$ 3.16   &   4.83  $\pm$ 3.18   &   4.45  $\pm$ 3.98  & 14.21  $\pm$11.84   &\bf2.3  $\pm$ 6.1  &  7.39  \\
    			Hip flexion MAE [deg]     &   3.92  $\pm$ 2.82   &   4.74  $\pm$ 3.45   &   4.26  $\pm$ 3.08  & 15.30  $\pm$11.38   &\bf2.6  $\pm$ 5.3  &  5.74  \\
    			Lean angle MAE [deg]      &   3.91  $\pm$ 2.60   &   3.68  $\pm$ 2.56   &   4.48  $\pm$ 4.61  &  8.09  $\pm$ 7.01   &\bf3.3  $\pm$ 3.3  &  n/a   \\
    			Fore/aft angle MAE [deg]  &   6.77  $\pm$ 5.20   &\bf5.75  $\pm$ 5.67   &   8.72  $\pm$ 8.42  & 12.26  $\pm$11.72   &  n/a              &  n/a   \\
    			Fore/aft distance MAE [m] &   0.07  $\pm$ 0.05   &   0.07  $\pm$ 0.06   &   0.09  $\pm$ 0.08  &  0.30  $\pm$ 0.38   &\bf0.03 $\pm$ 0.05 &  n/a   \\
    			\hline%
    		\end{tabular}%
    	}%
    	\end{center}%
    	\vspace*{-0.5cm}
    	\caption{%
    	    Comparison of our results in calibrated and uncalibrated cases to methods proposed by O\v{s}trek \etal (unpublished data) (C:~\textit{Monocular 3D pose estimation}) and Rhodin \etal~\cite{RhodinUnsupervisedSki} (D:~\textit{Semi-supervised}).
    	    As a baseline, we also provide results when not using a cosine basis and directly optimizing for 3D pose coordinates (Ours-A and Ours-B). In the uncalibrated case (Ours-B), we show the performance when optimizing camera rotations without enforcing Equation~\ref{eq:error_rot}.}
    	\label{tab:comparison}
    \end{table*}
    
    In Table~\ref{tab:comparison} we highlight the best results obtained from both the calibrated and uncalibrated cases. For those, the bundle adjustment used all 6 cameras and 2D pose estimates by the best performing dataset configuration that includes one additional \emph{MV-Ski} sequence. For the local metrics, like the centered MPJPE and the biomechanical variables, the two methods both yield comparable, high accuracies, with the uncalibrated method only slightly worse. Still, the latter is mostly well within the standard deviations of the calibrated methods. On the global metrics, like the MPJPE, CoM distances and speed, we see a larger discrepancy between the two methods, meaning that the poses found when estimating the camera rotations are locally accurate, but globally contain small deviations of the whole pose.
    
    As baselines, we also report all metrics on bundle adjustment approaches, where we don't use the cosine basis parametrization (Ours-A and Ours-B). In the uncalibrated case (Ours-B) we also directly optimize for the rotation matrices without getting rotation measures from the homography approach. As can be seen in Table~\ref{tab:comparison}, our new uncalibrated approach outperforms the baseline by an order of magnitude in all metrics, with only slightly lower gains in the calibrated case. Note that the bundle adjustment in Ours-B easily gets stuck in a local degenerate minimum. While this approach worked for previous work~\cite{Takahashi}, the high speeds, large capture volume and zoomed in cameras in the \emph{MV-Ski} dataset impede convergence to better solutions. Using background cues and the cosine basis provided a strong guide to avoid those local minima. The other main benefits in that case are the improved smoothness of the motion and speedup of the bundle adjustment. Using the cosine basis parametrization, we usually need only about half the number of iterations to reach a similar performance.
    
    An analysis showcasing the performance of both calibrated and uncalibrated cases, across dataset configurations and number of cameras used, can be found in the supplementary material. We see the largest gains in accuracy when going from two to three cameras. With more than that, reconstruction quality still improves, but with diminishing returns. 
    We also show how improved 2D detection quality directly translates into higher 3D triangulation accuracy.

\para{Comparison to existing methods.}
	In Table~\ref{tab:comparison} we compare our best results with the methods proposed by O\v{s}trek \etal (unpublished data) and Rhodin \etal~\cite{RhodinUnsupervisedSki}. O\v{s}trek \etal's method computes all biomechanical variables indirectly from images via monocular 3D pose estimation trained on the \emph{MV-Ski} dataset. Rhodin \etal estimate a monocular 3D pose using a semi-supervised method by constraining the model to predict the same pose in all views and needing only few labelled images.
	Both our best calibrated and uncalibrated methods perform only slightly worse than O\v{s}trek \etal (unpublished data), while yielding lower standard deviations and using only a single sequence from the \emph{MV-Ski} dataset. 
	Both our methods outperform the semi-supervised method by Rhodin \etal~\cite{RhodinUnsupervisedSki}.

\para{Velocity estimation.}
	In contrast to existing monocular approaches~\cite{RhodinUnsupervisedSki}, our method allows to estimate the athletes instantaneous velocity as the change in CoM position between two frames, even if camera rotations are unknown. Using calibrated cameras, we get mean absolute errors of 0.74 $\pm$ 1.76 m/s, while when estimating rotations, it rises to 1.87 $\pm$ 2.64 m/s. This is still relatively low, given the high speed of the professional athletes, that ranges between $15-20$ m/s in the test sequences and the large capture volume of more than $30-50$ m distance between cameras.
\section{Conclusion}
	We developed a practical method for reconstructing the global articulated 3D pose from bare videos taken from multiple rotating cameras.
    Our key contribution is joint optimization of 3D human pose and camera rotation by incorporating additional constraints from tracked features and resulting homographies.
    Our empirical evaluation shows that training 2D keypoint detection on the large \emph{SV-Ski} dataset and subsequent multi-view 3D reconstruction is as accurate or better as training 3D pose estimation directly on the available small-scale multi-view datasets~\cite{RhodinUnsupervisedSki}, while promising improved generalization capability to new scenes. The improvement brought about by our contributions are quantified in terms of widely used reconstruction metrics as well as biomechanical variables that are common for performance analysis of professional athletes. By contrast to monocular solutions, we are able to provide accurate global measurements without the need for cumbersome camera rotation calibration, which makes this method directly applicable for ski coaches.
    
    \paragraph{Acknowledgement.} This work was supported in part by the Swiss National Science Foundation.
\clearpage

{\small
\bibliographystyle{ieee}
\bibliography{references}
}

\clearpage
\onecolumn
\begin{appendices}
\label{appendix}

In this supplementary material, we analyze the performance gains of using different dataset configurations in both 2D and 3D pose estimation and show how various 3D metrics perform as a function of the number of cameras used. Further, we explain how those metrics were calculated and give training specifications for both the object detector and pose estimator networks.

\section{2D pose estimation performance as a function of dataset modifications}\label{ap:2dpose_plots}
\begin{figure}[!htb]
		\begin{center}
			\includegraphics[width=\linewidth]{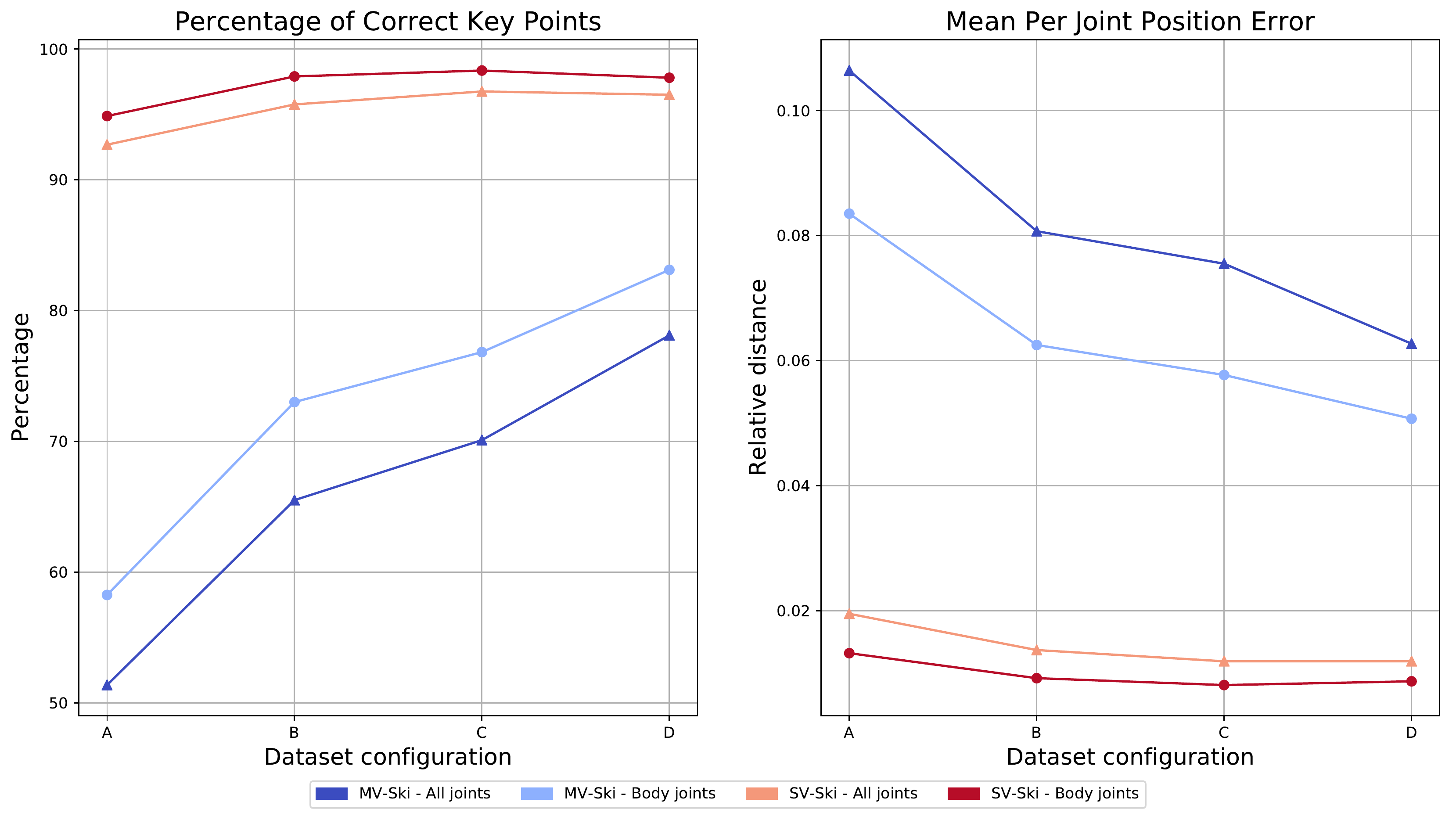}
		\end{center}
		\caption{Comparison of all four dataset configurations, showing the improvements each addition yields on unseen test sets for both the \emph{SV-Ski} and \emph{MV-Ski} datasets. A: \textit{Alpine dataset}, B: \textit{Alpine dataset augmented with calibration poles}, C: \textit{Transfer learning from MPII weights on augmented dataset}, D: \textit{Like previous, but with one additional \emph{MV-Ski} sequence from four camera angles}.}
		\label{fig:2dpose_results}
	\end{figure}

    We evaluate different OpenPose trainings on test sets of both the \emph{SV-Ski} and \emph{MV-Ski} datasets. We report metrics on both all the athlete's joints, as well as when excluding skis and poles.
	With our 2D object detector and pose estimator trained on mainly the new \emph{SV-Ski} dataset we see them under-performing on the \emph{MV-Ski} dataset because it contains challenging features, like the calibration poles. The improvements on the accuracy by a range of modifications are displayed graphically in Figure~\ref{fig:2dpose_results}. The addition of cut-out calibration poles onto training images improves accuracies by a large margin (Configuration A to B) for the \emph{MV-Ski} test set. By adding one sequence from the \emph{MV-Ski} dataset, we again see a considerable jump in performance on the \emph{MV-Ski} test set. Performance on the \emph{SV-Ski} is already very high and increases strongest when adding data augmentation, making it more robust.

\clearpage
\section{3D performance as a function of number of cameras and dataset modifications}\label{ap:camera_plots}

\begin{figure}[!htb]
		\begin{center}
			\includegraphics[width=0.69\linewidth]{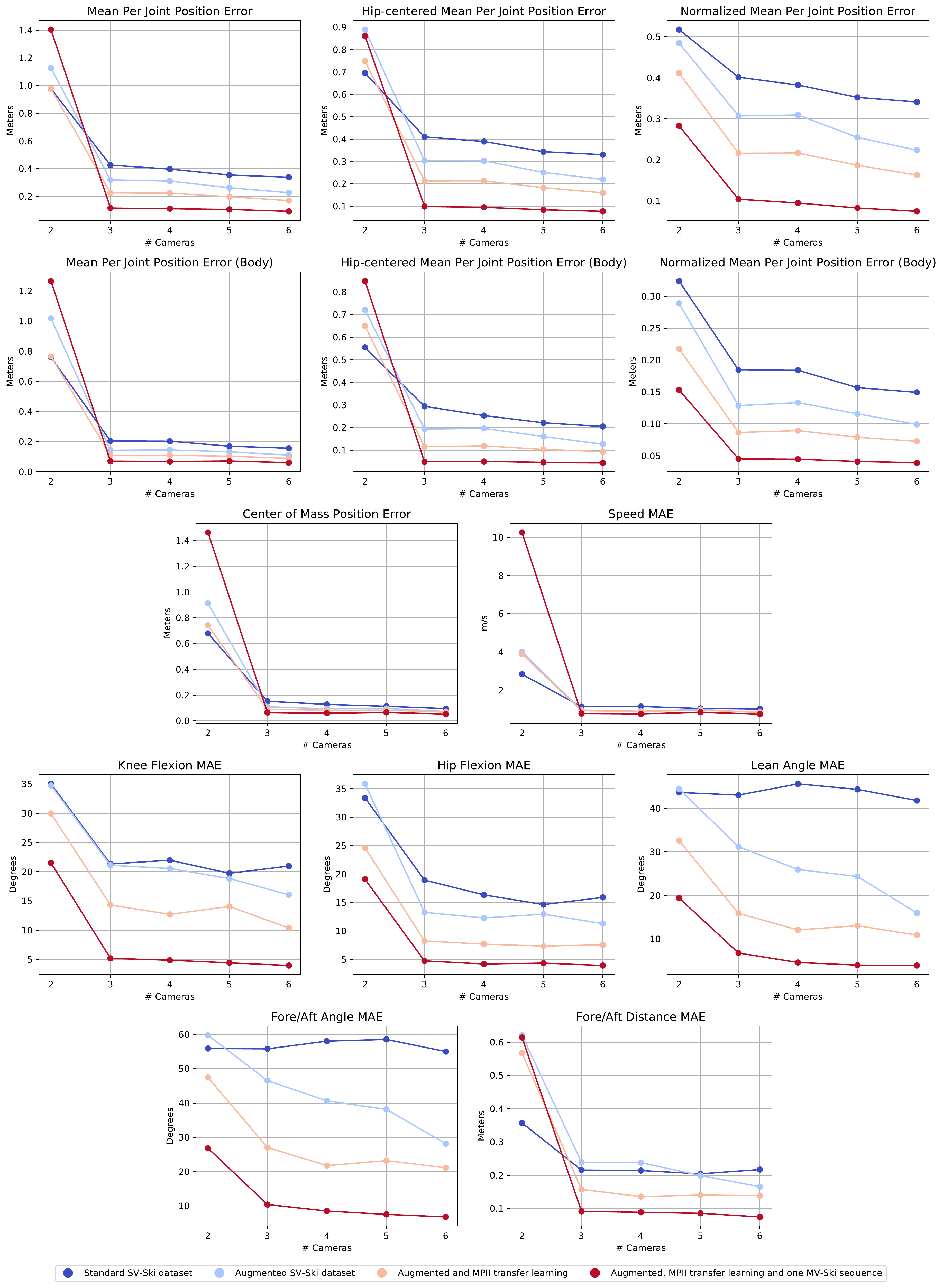}
		\end{center}
		\caption{With known rotation matrices: Comparison of all performance metrics for different number of cameras used in the bundle adjustment and four differently trained 2D detectors.}
		\label{fig:3D_results_known_rot}
	\end{figure}

\begin{figure}[!htb]
		\begin{center}
			\includegraphics[width=0.69\linewidth]{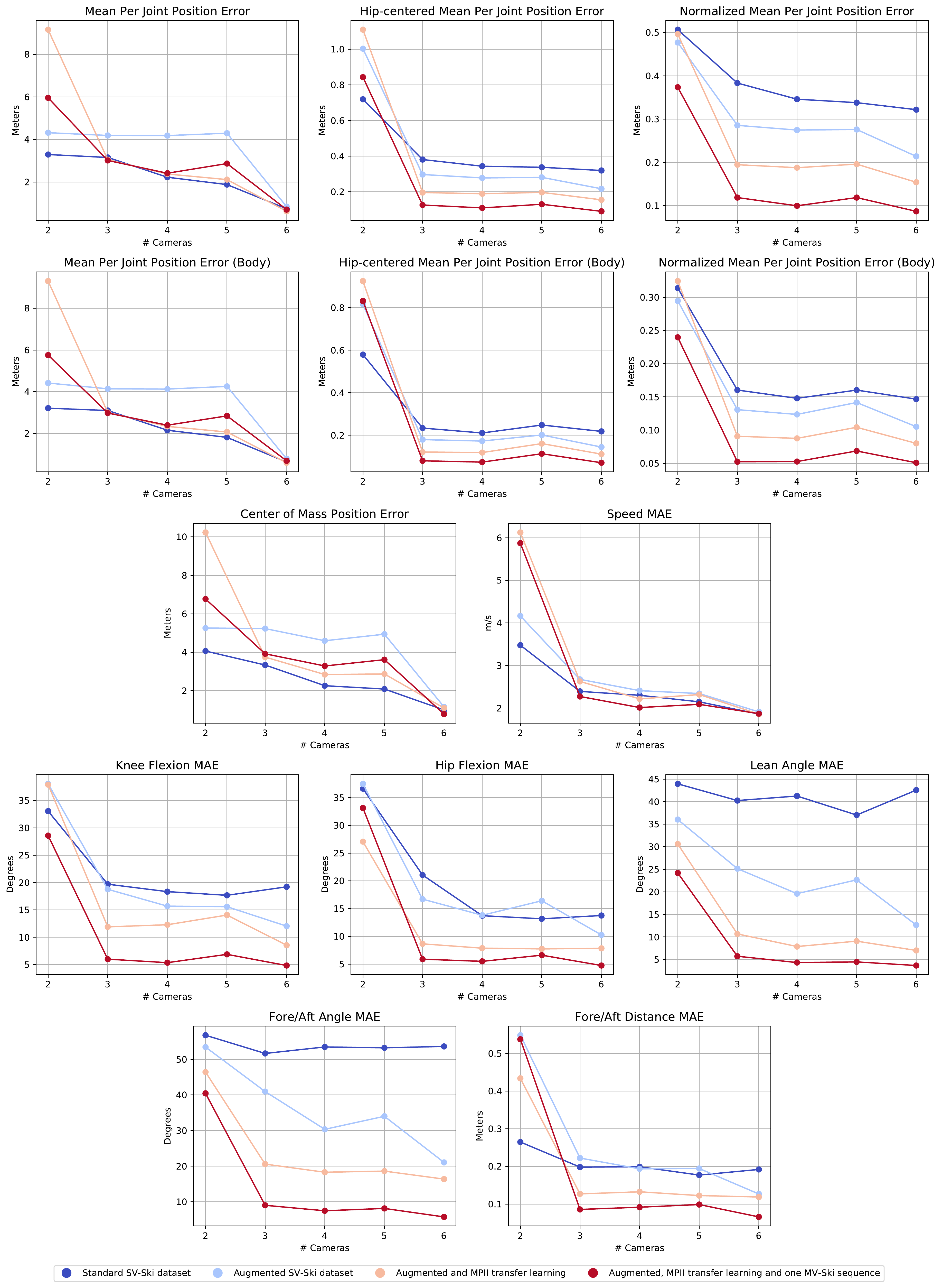}
		\end{center}
		\caption{Estimating rotation matrices: Comparison of all performance metrics for different number of cameras used in the bundle adjustment and four differently trained 2D detectors.}
		\label{fig:3D_results_est_rot}
	\end{figure}

    Figures~\ref{fig:3D_results_known_rot} and~\ref{fig:3D_results_est_rot} show various performance metrics extracted from comparing the estimated 3D poses with the ground truth, in the cases where camera rotations are known and where we estimate them. For each metric, we show the performance across dataset configurations and number of cameras used. All metrics were averaged over two unseen test sequences.
    
    In all cases, the methods utilizing all 6 cameras with the OpenPose weights that were trained on one additional \emph{MV-Ski} sequence consistently performed the best. We see the largest jump in performance when taking 3 instead of 2 cameras. In the calibrated case, using more cameras still improves all metrics, but with diminishing returns. When estimating camera rotations, local metrics (like the biomechanical variables or hip-centered MPJPE) see no big improvements using more than 3 cameras, whereas the global metrics (like the MPJPE or CoM) benefit from each additional camera.
    
\clearpage
\section{Metrics}\label{ap:metrics}
	We compute performance scores using both general metrics using merely the locations of estimated and ground truth joints, as well as more specific errors between estimated and ground truth biomechanical variables.

\subsection{General performance metrics}\label{ap:general_perf}
\paragraph{Percentage of Correct Key Points (PCK)}\label{ap:pck}
	The PCK is defined as the percentage of estimated 2D points $\bm{\hat{p}}_j^f$ detected within radius $r^f$ of the ground truth $\bm{p}_j^f$ as
	\begin{equation}
	PCK = \frac{100}{N_F N_J} \sum_{f=1}^{N_F} \sum_{j=1}^{N_J} \mathds{1} (\norm{\bm{\hat{p}}_j^f - \bm{p}_j^f}_2 \leq r^f) \;.
	\end{equation}
	We take the radius $r^f$ as the ground truth head-to-neck distance
	\begin{equation}
	r^f = \norm{\bm{p}_{head}^f - \bm{p}_{neck}^f}_2 \;.
	\end{equation}
	In the \emph{MV-Ski} dataset, the head joint is not at the top but in the middle, therefore we use the double distance $r^f$ in that case.

\paragraph{Average distances}\label{ap:mpjpe}
	In both 2D and 3D cases, the Mean Per Joint Position Error (MPJPE) is defined as
	\begin{equation}
	MPJPE = \frac{1}{N_F N_J} \sum_{f=1}^{N_F} \sum_{j=1}^{N_J} \norm{\bm{\hat{p}}_j^f - \bm{p}_j^f}_2
	\end{equation}
	over all $N_F$ frames and $N_J$ joints in the dataset. In the 3D case only, we calculate the hip-centered MPJPE, where we subtracted the center-hip position from every joint. We also compute the Normalized MPJPE (NMPJPE)~\cite{RhodinUnsupervisedSki}, which we calculate by the usual MPJPE metric between the \emph{hip-centered} ground truth and normalized \emph{hip-centered} poses, given by
	\begin{equation}
	\bm{\hat{p}}_{norm}^f =\frac{\bm{\hat{p}}^f \cdot \bm{p}^f}{\bm{\hat{p}}^f \cdot \bm{\hat{p}}^f} \bm{\hat{p}}^f \;,
	\end{equation}
	where pose vectors are flattened for the dot products.

\paragraph{Center of Mass (CoM)}\label{ap:com}
	The CoM of a person is calculated as a weighted sum of limb centers
	\begin{equation}
	CoM = \frac{1}{M} \sum_{(i,j) \in Limbs} m_{(i,j)} \frac{\bm{P}_i + \bm{P}_j}{2} \;,
	\end{equation}
	where $m_{(i,j)}$ is the mass of the limb connecting joints $i$ and $j$, and $M = \sum_{(i,j) \in Limbs} m_{(i,j)}$.
	
	The relative weights, adapted from O\v{s}trek \etal (unpublished data), are the following for both sides of the body: 
	\textit{(head)}: 0.065, \textit{(shoulder, hip)}: 0.1835, \textit{(shoulder, elbow)}: 0.023, \textit{(elbow, hand)}: 0.014, \textit{(hand)}: 0.006, \textit{(hip, knee)}: 0.119, \textit{(knee, ankle)}: 0.038, \textit{(toes, heel)}: 0.038, \textit{(ski tip, ski tail)}: 0.043, \textit{(hand, ski pole basket)}: 0.003 .

\paragraph{Speed}\label{ap:speed}
	Knowing the global 3D positions of the skier's COM and the frame rate of the videos we can compute their speed $v_{CoM}^f$ at every time step $f$ using
	\begin{equation}
	v_{CoM}^f = \norm{CoM^f - CoM^{f+1}} FPS \;.
	\end{equation}
	All \emph{MV-Ski} Videos feature a frame rate of 50 Frames Per Second (FPS). For calculating the mean absolute error between the ground truth and predicted speeds, the predicted CoM's have been smoothed temporally by a Gaussian filter with $\sigma = 1.5$.

\subsection{Biomechanical variables}\label{ap:biomech_vars}
	More specific to alpine skiing, there are several biomechanical variables that are used for injury prevention or training purposes. We mainly focus on knee flexion, hip flexion, lean angle and fore/aft angle/distance. All variables are calculated from the global 3D pose $\bm{P}$.

\paragraph{Knee flexion}\label{ap:knee_flexion}
	The knee flexion is the angle measuring how much the knee is bent. It is calculated as the angle between the two vectors
	\begin{align}
	\bm{v}_{thigh} &= \bm{P}_{hip} - \bm{P}_{knee} \\
	\bm{v}_{crus} &= \bm{P}_{ankle} - \bm{P}_{knee}
	\end{align}
	for the right and left legs in the same way. The angle between those vectors is given by
	\begin{equation}
	\theta_{knee} = \arccos \frac{\bm{v}_{thigh} \cdot \bm{v}_{crus}}{\norm{\bm{v}_{thigh}}  \norm{\bm{v}_{crus}}} \;.
	\end{equation}

\paragraph{Hip flexion}\label{ap:hip_flexion}
	The hip flexion is the angle between the spine and right or left thigh. Those vectors are defined as
	\begin{align}
	\bm{v}_{spine} &= \bm{P}_{neck} - \bm{P}_{center\_hip} \\
	\bm{v}_{thigh} &= \bm{P}_{knee} - \bm{P}_{hip}
	\end{align}
	and again we calculate the angle between them by
	\begin{equation}
	\theta_{hip} = \arccos \frac{\bm{v}_{spine} \cdot \bm{v}_{thigh}}{\norm{\bm{v}_{spine}}  \norm{\bm{v}_{thigh}}} \;.
	\end{equation}

\paragraph{Lean angle}\label{ap:lean_angle}
	The lean angle is the sideways inclination between the CoM and the left or right middle of the skis. It is usually calculated on the outside ski, which can be reasonably well figured out by taking the leg with the larger knee flexion value. The following equations are only given for the right side, but symmetry applies. First we define the local coordinate vectors $x'$, $y'$ and $z'$, fixed at the center of both ankles:
	\begin{align}
	\label{eq:xyz_ankle_a}
	\bm{x}' &= \bm{P}_{right\_ski\_tip} - \bm{P}_{right\_ski\_tail} \\
	\bm{y} & = \bm{P}_{left\_ankle} - \bm{P}_{right\_ankle} \\
	\bm{z}' &= \bm{x}' \times \bm{y} \\
	\label{eq:xyz_ankle_b}
	\bm{y}' &= \bm{z}' \times \bm{x}' \;,
	\end{align}
	where $\bm{x}'$ and $\bm{y}$ have been normalized after calculation.
	Furthermore, we define the vector $\bm{C}$ between the center of both ankles
	\begin{equation}
	\bm{P}_{center\_ankle} =  \frac{\bm{P}_{left\_ankle} + \bm{P}_{right\_ankle}}{2}
	\end{equation}
	and the CoM as
	\begin{equation}
	\bm{C} = CoM - \bm{P}_{center\_ankle} \;.
	\end{equation}
	Now we need to project $\bm{C}$ onto the $\bm{y}'\bm{z}'$ plane as
	\begin{equation}
	\pi_{\bm{y}'\bm{z}'}(\bm{C}) = \bm{C} - (\bm{C} \cdot \bm{x}') \cdot \bm{x}'
	\end{equation}
	and can then compute the lean angle as the angle $\lambda_{lean}$ between the normalized projection and the $\bm{z}'$ axis using
	\begin{equation}
	\lambda_{lean} = \arccos (\bm{z}' \cdot \pi_{\bm{y}'\bm{z}'}(\bm{C})) \;.
	\end{equation}

\paragraph{Fore/aft angle and distance}\label{ap:fore_aft}
	The fore/aft angle measures how much the skier leans to the front or back, instead of the side. To calculate it, we use the same $x'$, $y'$ and $z'$ vectors as calculated in equations \ref{eq:xyz_ankle_a} - \ref{eq:xyz_ankle_b}, but centered at the outward ski ankle. We define the vector $\bm{C}$ now as
	\begin{equation}
	\bm{C} = CoM - \bm{P}_{right\_ankle} \;.
	\end{equation}
	and project it onto the $\bm{x}'\bm{z}'$ plane as
	\begin{equation}
	\pi_{\bm{x}'\bm{z}'}(\bm{C}) = \bm{C} - (\bm{C} \cdot \bm{y}') \cdot \bm{y}' \;.
	\end{equation}
	Then we can compute the fore/aft angle $\lambda_{fore/aft}$ between the normalized projection and the $\bm{z}'$ axis using
	\begin{equation}
	\lambda_{fore/aft} = \arccos (\bm{z}' \cdot \pi_{\bm{y}'\bm{z}'}(\bm{C})) \;.
	\end{equation}
	The last metrics we compute is the fore/aft distance $d_{fore/aft}$, defined as
	\begin{equation}
	d_{fore/aft} = \sin(\lambda_{fore/aft}) \cdot \norm{\bm{C}} \;.
	\end{equation}

\section{Training specifications}\label{ap:train_specs}
This section provides more information on how the SSD and OpenPose networks were trained.

\subsection{SSD training specifications}\label{ap:train_specs_ssd}
    We trained a PyTorch implementation of the SSD network~\cite{ssd-pytorch} on the newly annotated \emph{SV-Ski} dataset for 2000 iterations, using a batch size of 32 images. The network was initialized with pretrained VGG 16 weights. The optimizer used was Stochastic Gradient Descent (SGD) with learning rate 0.001, momentum 0.9 and weight decay 0.0005. The learning rate was scaled by 0.1 at iterations 1000 and 1500. As this PyTorch implementation was made for multiple object classes we trained it with one athlete class and one unused dummy class.

\subsection{OpenPose training specifications}\label{ap:train_specs_openpose}
    For training OpenPose, all images were resized to 736~x~368 pixels. In addition to the augmentation with \emph{MV-Ski} calibration poles, for data augmentation each of the following transformations was applied independently with a probability of 0.5:

	\begin{itemize}
		\item Adjusting the image gamma value uniformly by a factor of $\mathcal{U}(0.5,1.5)$.
		\item Shifting the image hue uniformly by $\mathcal{U}(-15,15)$ \degree.
		\item Rotating the image uniformly by $\mathcal{U}(-40,40)$ \degree.
		\item Mirroring the image horizontally.
	\end{itemize}

	In addition, all training images were randomly cropped around the ground truth pose of the skier, such that the network always received poses of roughly the same scale. During test time, no data augmentation is applied besides resizing the imaged to the input resolution.

	Training was done in batch sizes of 8 over 200 epochs using the Adam optimizer with learning rate 0.00004, momentum 0.9 and weight decay 0.0005. In all cases, the first OpenPose stage block was initialized with pretrained VGG 19 weights, while all other stages were either randomly initialized or using weights pretrained on the MPII dataset.

	During test time the OpenPose model that yielded the lowest validation error during all training epochs was chosen. The outputs from the SSD network were resized to 368~x~368 pixels and run through OpenPose several times scaled by 0.5, 0.75, 1, 1.25, 1.5 and 2 for being able to detect a wider range of SSD outputs. Results from all scales were averaged to generate the final heatmaps and PAFs.

\end{appendices}

\end{document}